\documentclass[runningheads]{llncs}
\usepackage{amssymb} %
\usepackage{amsmath} 
\usepackage{bm}

\newcommand{\rcoord}{\mathbf{r}}
\newcommand{\kcoord}{\mathbf{k}}
\newcommand{\Peq}{T_{\text{R}} / T_{\text{10}}(\rcoord)}
\newcommand{\Qeq}{r_1 C(\rcoord,t) T_{\text{R}}}

\usepackage{hyperref}
\hypersetup{colorlinks   = true, 
linkcolor    = red, 
citecolor   = blue  
}

\usepackage{graphicx} 
\usepackage{epstopdf}
\usepackage{floatrow}

\usepackage{algpseudocode}
\usepackage{algorithm}

\usepackage{multirow}
\newif\ifalgo
\algofalse
\begin{document}

\setlength{\abovecaptionskip}{15pt plus 3pt minus 2pt}

\setlength{\abovedisplayskip}{3pt}
\setlength{\belowdisplayskip}{3pt}

\title{Direct Estimation of Pharmacokinetic Parameters from DCE-MRI using Deep CNN with Forward Physical Model Loss}


\author{Cagdas Ulas$^{1}$, Giles Tetteh$^{1}$, Michael J. Thrippleton$^{2}$, Paul A. Armitage$^{4}$, Stephen D. Makin$^{2}$, Joanna M. Wardlaw$^{2}$, Mike E. Davies$^{3}$, Bjoern H. Menze$^{1}$}
\institute{
$^1$ Department of Computer Science, Technische Universit\"{a}t M\"{u}nchen, Germany\\
$^2$ Department of Neuroimaging Sciences, University of Edinburgh, UK\\
$^3$ Institute for Digital Communications, University of Edinburgh, UK \\
$^4$Department of Cardiovascular Sciences,University of Sheffield, UK}

\titlerunning{Direct Estimation of PK Parameters of DCE-MRI using Deep CNN}

\authorrunning{Ulas et al.}



\maketitle

\begin{abstract}
Dynamic contrast-enhanced (DCE) MRI is an evolving imaging technique that provides a quantitative measure of pharmacokinetic (PK) parameters in body tissues, in which series of $T_1$-weighted images are collected following the administration of a paramagnetic contrast agent. Unfortunately, in many applications, conventional clinical DCE-MRI suffers from low spatiotemporal resolution and insufficient volume coverage. In this paper, we propose a novel deep learning based approach to directly estimate the PK parameters from undersampled DCE-MRI data. Specifically, we design a custom loss function where we incorporate a forward physical model that relates the PK parameters to corrupted image-time series obtained due to subsampling in k-space. This allows the network to directly exploit the knowledge of true contrast agent kinetics in the training phase, and hence provide more accurate restoration of PK parameters. Experiments on clinical brain DCE datasets demonstrate the efficacy of our approach in terms of fidelity of PK parameter reconstruction and significantly faster parameter inference compared to a model-based iterative reconstruction method.

\end{abstract}



\section{Introduction} \label{sec:Intro}

Dynamic contrast-enhanced (DCE) MRI involves the administration of a $T_1$-shortening Gadolinium-based contrast agent (CA), followed by the acquisition of successive $T_1$-weighted images as the contrast bolus enters and subsequently leaves the organ \cite{Sourbron2013}. In DCE-MRI, changes in CA concentration are derived from changes in signal intensity over time, then regressed to estimate pharmacokinetic (PK) parameters related to vascular permeability and tissue perfusion \cite{Lebel2014}.
Since perfusion and permeability are typically affected in the presence of vascular and cellular irregularities, DCE imaging has been considered as a promising tool for clinical diagnostics of brain tumours, multiple sclerosis lesions, and neurological disorders where disruption of blood-brain barrier (BBB) occurs. \cite{Oconnor2012,Heye2016}.

Despite its effectiveness in quantitative assessment of microvascular properties, conventional DCE-MRI is challenged by suboptimal image acquisition that severely restricts the spatiotemporal resolution and volume coverage \cite{Guo2016,Guo2017}. The shortest possible scanning time often leads to limited spatial resolution hampering detection of small image features and accurate tumor boundaries. Low temporal resolution hinders accurate fitting of PK parameters. Furthermore, volume coverage is usually inadequate to cover the known pathology, for instance in the case multiple metastatic lesions \cite{Guo2017}. Facing such severe constraints, DCE imaging can significantly benefit from undersampled acquisitions.

So far, existing works in
\cite{Lebel2014,Guo2016,Zhang2015} have proposed  compressed sensing and parallel imaging based reconstruction schemes to accelerate DCE-MRI acquisitions, mainly targeting to achieve better spatial resolution and volume coverage while retaining the same temporal resolution. These methods are referred to as indirect methods \cite{Guo2017} because they are based on the reconstruction of dynamic DCE image series first, followed by a separate step for fitting the PK parameters on a voxel-by-voxel level using a tracer kinetic model \cite{Sourbron2013}. More recently, a model-based direct reconstruction model \cite{Guo2017} has been proposed to directly estimate PK parameters from undersampled (k,t) space data. The direct reconstruction method generally poses the estimation of PK maps as an error minimization problem. This approach has been shown to produce superior PK parameter maps and allows for higher acceleration compared to indirect methods. However, the main drawback of this method is that parameter reconstruction of an entire volume requires considerably high computation time.

Motivated by the recent advances of deep learning in medical imaging, in this paper, we present a novel deep learning based approach to directly estimate PK parameters from undersampled DCE-MRI data. First, our proposed network takes the corrupted image-time series as input and \textit{residual} parameter maps, which represent deviations from a kinetic model fitting on fully-sampled image-time series, as output, and aims at learning a nonlinear mapping between them. 
Our motivation for learning the \textit{residual} PK maps is based on the observation that residual maps are more sparse and topologically less complex compared to target parameter maps. Second, we propose the \textit{forward physical model loss}, a custom loss function in which we exploit the physical relation between true contrast agent kinetics and measured time-resolved DCE signals when training our network.
Third, we validate our method experimentally on human \textit{in vivo} brain DCE-MRI dataset. 
We demonstrate the superior performance of our method in terms of parameter reconstruction accuracy and significantly faster estimation of parameters during testing, taking approximately 1.5 seconds on an entire 3D test volume. To the best of our knowledge, we present the first work leveraging the machine learning algorithms  -- specifically deep learning -- to directly estimate PK parameters  from undersampled DCE-MRI time-series.


\section{Methods} \label{sec:Methods}
We treat the parameter inference from undersampled data in DCE imaging as a mapping problem between the corrupted intensity-time series and \textit{residual} parameter maps where the underlying mapping is learned using deep convolutional neural networks (CNNs). We provide a summary of general tracer kinetic models applied in DCE-MRI in Sec. \ref{sec:DCE-MRI}, formulate the forward physical model relating the PK parameters to undersampled data in Sec. \ref{sec:pyhsical-model}, finally describe our proposed deep learning methodology for PK parameter inference in Sec. \ref{sec:deep-cnn}.

\subsection{Tracer Kinetic Modeling in DCE-MRI} \label{sec:DCE-MRI}
\vspace{-1mm}
Tracer kinetic modeling aims at providing a link between the tissue signal enhancement and the physiological or so-called pharmacokinetic parameters, including the fractional plasma volume ($v_{\text{p}}$), the fractional interstitial volume ($v_{\text{e}}$), and the volume transfer rate ($K^\text{trans}$) at which contrast agent (CA) is delivered to the extravascular extracellular space (EES). 
\begin{figure}[t!]
 \centering
 \includegraphics[width=0.92\columnwidth]{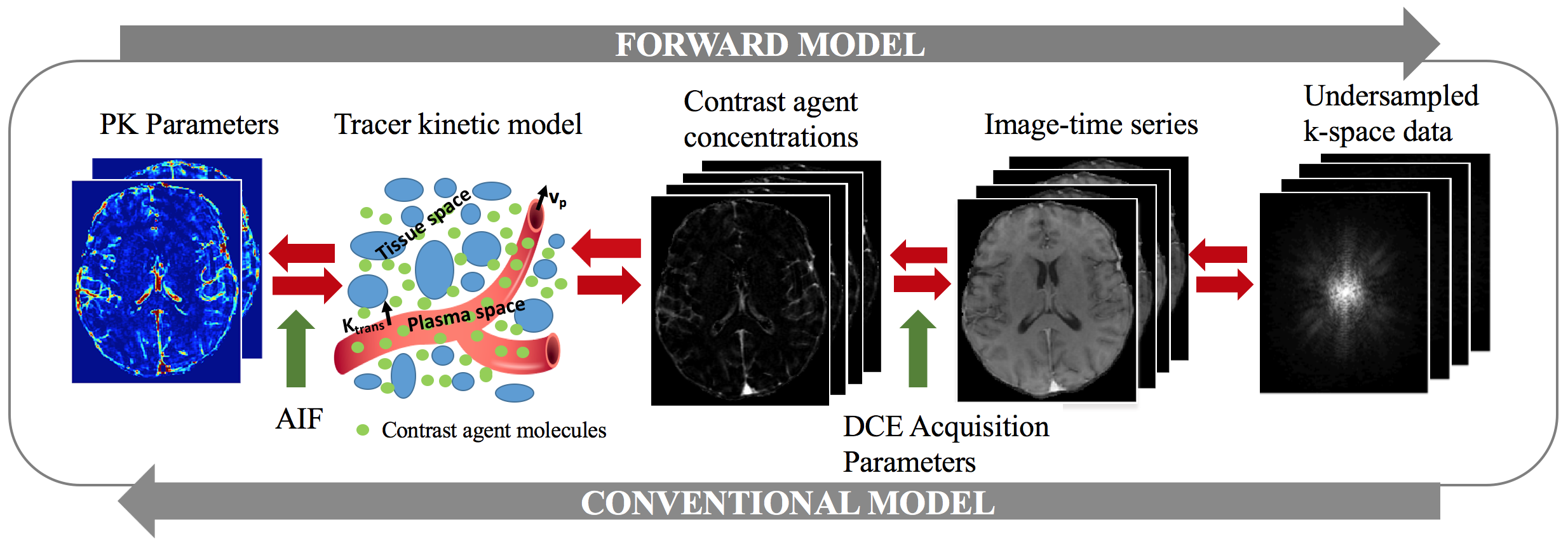}
 \caption{Computational steps in the forward model and the conventional pipeline of PK parameter estimation in DCE-MRI. \label{fig:DCEmodel}}
\end{figure}
One of the well-established tracer kinetic models is known as Patlak model \cite{Patlak1983}. This model describes a highly perfused two compartment tissue, ignoring backflux from the EES into the blood plasma compartment. The CA concentration in the tissues is determined by,
\begin{equation}
C(\rcoord,t) = v_p(\rcoord)C_p(t) + K^{\text{trans}} (\rcoord)\int_0^{t} C_p(\tau) d\tau,
\label{Patlak-model-eqn}
\end{equation} 
where $\rcoord \in (x,y,z)$ represent image domain spatial coordinates, $C(\rcoord,t)$ is the CA concentration over time, and $C_p(t)$ denotes the arterial input function (AIF) which is usually measured from voxels in a feeding artery.

In this work, we specifically employ the Patlak model for tracer pharmacokinetic modeling and estimation of ground truth tissue parameters. This model is a perfect match for our DCE dataset because it is often applied when the temporal resolution is too low to measure the cerebral blood flow, and it has been commonly used to measure the BBB leakage with DCE-MRI in acute brain stroke and dementia \cite{Heye2016,Sourbron2013}. An attractive feature of Patlak model is that the model equation in (\ref{Patlak-model-eqn}) can be linearized and fitted using linear least squares which has a closed-form solution, hence parameter estimation is fast \cite{Sourbron2013}.

\subsection{Forward Physical Model: From PK Parameters to Undersampled Data}
\label{sec:pyhsical-model}

Figure~\ref{fig:DCEmodel} depicts the conventional and forward model approaches relating the PK parameter estimation to undersampled or fully-sampled k-space data, and vice versa. 
For direct estimation of PK parameters from the measured k-space data, as proposed in \cite{Fang2016,Guo2017}, a forward model can be formulated by inverting the steps in the conventional model as follows:
\begin{enumerate}
\item Given the sets of PK parameter pairs ($K^{\text{trans}} (\rcoord), v_{\text{p}} (\rcoord)$) and arterial input function $C_p(t)$,  CA concentration curves over time $C(\rcoord,t)$ are estimated using the Patlak model equation in (\ref{Patlak-model-eqn}).
\item Dynamic DCE image series $S(\rcoord,t)$ are converted to $C(\rcoord,t)$ through the steady-state spoiled gradient echo (SGPR) signal equation \cite{Guo2017}, given by
\begin{equation}
S(\rcoord,t) = \frac{M_{\text{0}}(\rcoord)\text{sin}\alpha(1-e^{-(K+L)})}{1-\text{cos}\alpha e^{-(K+L)}} +\left(S(\rcoord,0) - \frac{M_{\text{0}}(\rcoord)\text{sin}\alpha(1-e^{-K})}{1-\text{cos}\alpha e^{-K}}\right)
\label{signal-eqn}
\end{equation}
where $K = \Peq$, $L= \Qeq$, $T_{\text{R}}$ is the repetition time, $\alpha$ is the flip angle, $r_1$ is the contrast agent relaxivity taken as 4.2  $\text{s}^{-1}\text{mM}^{-1}$, $S(\rcoord,0)$ is the baseline (pre-contrast) image intensity, and  $T_{\text{10}}(\rcoord)$  and $M_{\text{0}}(\rcoord)$ are respectively the $T_{\text{1}}$ relaxation and equilibrium longitudinal magnetization that are calculated from a pre-contrast $T_{\text{1}}$  mapping acquisition.

\item The undersampled raw (k,t)-space data $S(\kcoord,t)$ can be related to $S(\rcoord,t)$ for a single-coil data by an undersampling fast Fourier transform (FFT), $F_u$,
\begin{equation}
S(\kcoord,t) = F_uS(\rcoord,t),
\label{fu-eqn}
\end{equation}
where $\kcoord \in (k_x,k_y,k_z)$ represents k-space coordinates.
\end{enumerate}
\vspace{-1mm}
By simply integrating the three computation steps in (\ref{Patlak-model-eqn}-\ref{fu-eqn}), we can form a single function $f_m$ modeling the signal evolution in (k-t) space
given the PK maps $\theta = \{K^{\text{trans}} (\rcoord), v_{\text{p}} (\rcoord) \}$, as $S(\kcoord,t)  = f_m(\theta; \bm{\xi})$, where $\bm{\xi}$ denotes all the predetermined acquisition parameters as mentioned above.

Given the undersampled (k,t)-space data $S(\kcoord,t)$, the corrupted image series $S_u(\rcoord,t)$ can be obtained by applying IFFT to $S(\kcoord,t)$, i.e. $S_u(\rcoord,t)  = F_u^{\mathsf{T}}S(\kcoord,t)$. We further define a new function $\boldsymbol{\tilde{f}_m}$ that integrates only the first two computation steps (\ref{Patlak-model-eqn}-\ref{signal-eqn}) to compute the dynamic DCE image series. We will incorporate $\boldsymbol{\tilde{f}_m}$ in our custom loss function that will be explained in the following section.

\begin{figure}[t!]
 \centering
 \includegraphics[width=0.85\columnwidth]{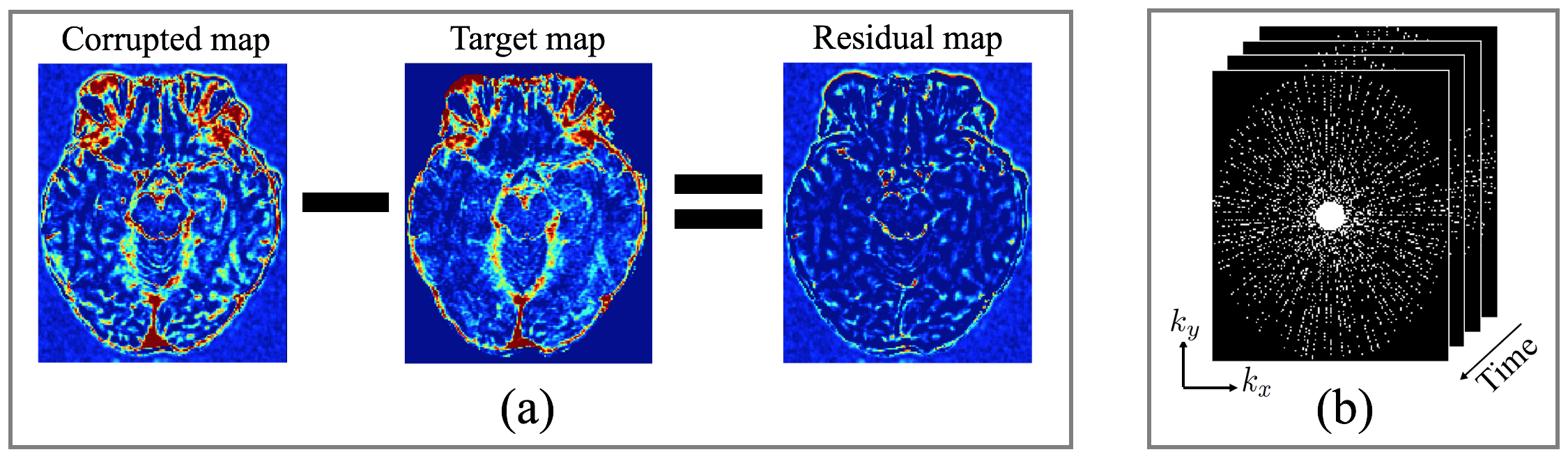}
 \caption{(a) The relation between a corrupted ($\theta_u$), target ($\theta_t$) and residual ($\theta_r$) PK maps, (b) Exemplary golden-angle sampling scheme in the $k_x$-$k_y$ plane through time. \label{fig:ResidualAndMask}}
\end{figure}

\subsection{PK Parameter Inference via Forward Physical Model Loss}
\label{sec:deep-cnn}

\subsubsection{Formulation.}
We hypothesize that a direct inversion between corrupted PK parameter maps $\theta_u$ and $S_u(\rcoord,t)$ is available through forward model, i.e., $S_u(\rcoord,t)  = \boldsymbol{\tilde{f}_m}(\theta_u)$. However, this cannot provide yet sufficiently accurate estimate of target parameter maps $\theta_t$ obtained from fully-sampled data $S(\rcoord,t)$. To this end, we estimate a correction or residual map $\theta_r$ from the available signal $S_u(\rcoord,t)$ satisfying $\theta_r= \theta_u - \theta_t$. As shown in Fig.~\ref{fig:ResidualAndMask}-(a), we observe that \textit{residual} PK maps involve more sparse representations and exhibit spatially less varying structures inside the brain. The task of learning a residual mapping was shown to be much easier and effective than the original mapping \cite{Zhang2017}. Following the same approach, we adopt the residual learning strategy using deep CNNs. Our CNN is trained to learn a mapping between $S_u(\rcoord,t)$ and $\theta_r$ to output an estimate of residual maps $\tilde{\theta}_r$; $\tilde{\theta}_r = \mathcal{R}(S_u(\rcoord,t) |\mathbf{\Theta})$, where $\mathcal{R}$ represents the forward mapping of the CNN parameterised by $\mathbf{\Theta}$. The final parameter estimate is obtained via $\tilde{\theta}_t = \theta_u - \tilde{\theta}_r$.

\subsubsection{Loss Function.}
We simultaneously seek the signal belonging to the corrected model estimates to be sufficiently close to true signal, i.e., $\boldsymbol{\tilde{f}_m}(\tilde{\theta}_t) \approx S(\rcoord,t)$. Therefore, we design a custom loss function which requires solving the forward model in every iteration of the network training. We refer the resulting loss as \textit{forward physical model loss}. Given a set of training samples $\mathcal{D}$ of input-output pairs ($S_u(\rcoord,t), \theta_r$), we train a CNN model that minimizes the following loss,
\begin{equation}
 \mathcal{L}(\mathbf{\Theta}) = \sum_{(S_u(\rcoord,t), \theta_r) \in \mathcal{D}} \lambda\|\theta_r - \tilde{\theta}_r\|_2^2  \enskip + \enskip (1-\lambda)\|S(\rcoord,t)- \boldsymbol{\tilde{f}_m}(\theta_u - \tilde{\theta}_r;\bm{\xi}) \|_2^2,
 \label{loss-eqn}
\end{equation}
where $\lambda$ is a regularization parameter balancing the trade-off between the fidelity of the parameter and signal reconstruction. We emphasize that the second term in (\ref{loss-eqn}) allows the network to intrinsically exploit the underlying contrast agent kinetics in training phase. %

\subsubsection{Network Architecture.} 
\begin{figure}[t!]
 \centering
 \includegraphics[width=0.9\columnwidth]{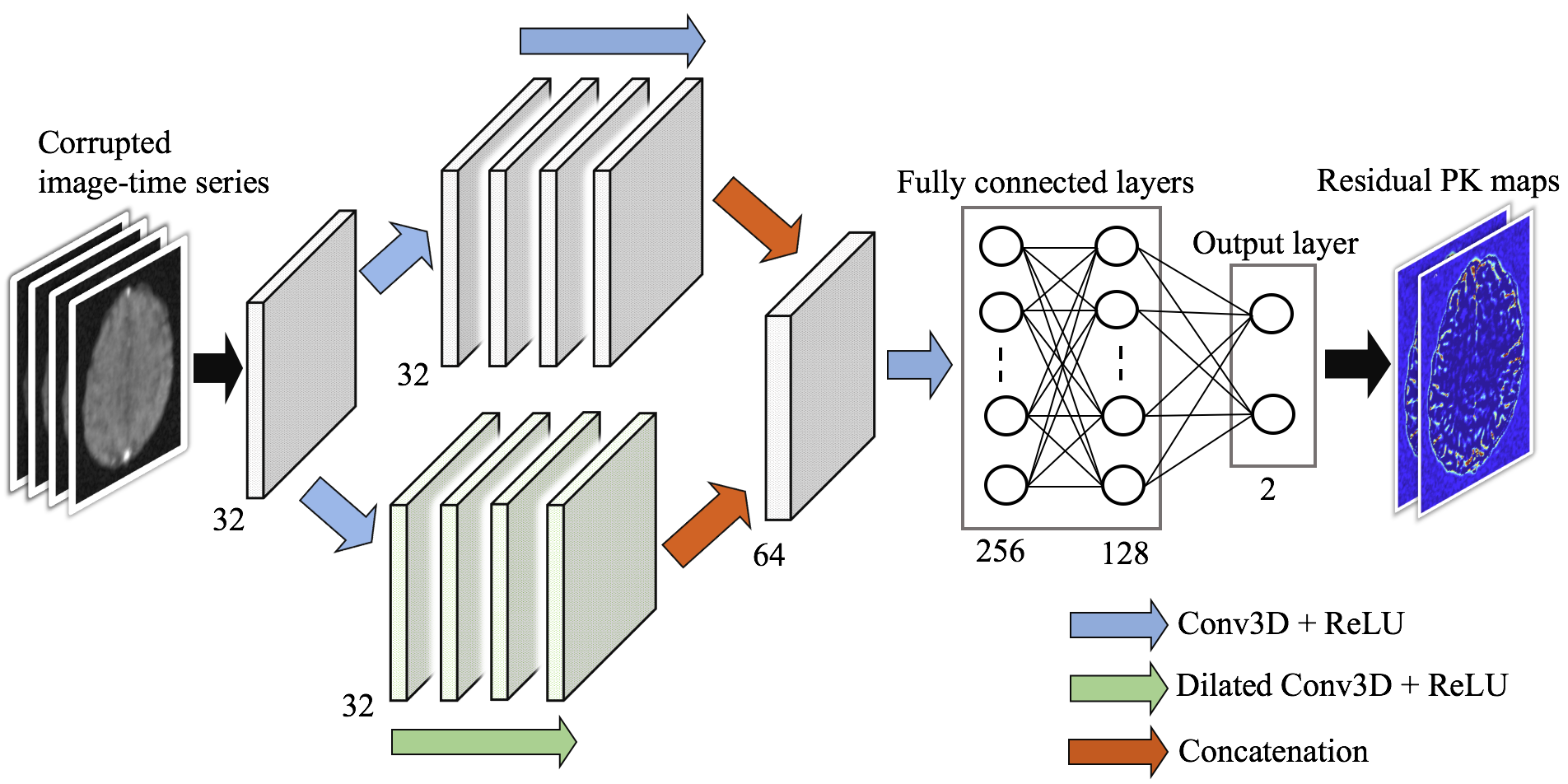}
 \caption{The network architecture used for the estimation of residual PK maps. The number of filters and output nodes are provided at the bottom of each layer.  \label{fig:NetworkArchitecture}}
\end{figure}

Figure~\ref{fig:NetworkArchitecture} illustrates our network architecture. The network takes a $4\text{D}$ image-time series as input, where time frames are stacked as input channels. The first convolutional layer applies $3\text{D}$ filters to each channel individually to extract low-level temporal features which are aggregated over frames via learned filter weights to produce a single output per voxel. Following the first layer, inspired by the work on brain segmentation \cite{Kamnitsas2017}, our network consists of parallel dual pathways to efficiently capture multi-scale information. The local pathway at the top focuses on extracting details from the local vicinity while the global pathway at the bottom is designed to incorporate more contextual global information. The global pathway consists of $4$ dilated convolutional layers with dilation factors of $2,4,8,16$, implying increased receptive field sizes. The filter size of each convolutional layer including dilated convolutions is $\mathrm{3\times3\times3}$, and the rectified linear units (ReLU) activation is applied after each convolution. Local and global pathways are then concatenated to form a multi-scale feature set. Following this, $2$ fully-connected layers are used to determine the best possible feature combination that can accurately map the input to output of the network. Finally, the last layer outputs the estimated residual maps.

\section{Experiments and Results}

\subsubsection{Datasets.}
We perform experiments on fully-sampled DCE-MRI datasets acquired from three mild ischaemic stroke patients. DCE image series were acquired using a 1.5T clinical scanner with a 3D T1W spoiled gradient echo sequence (TR/TE = $8.24/3.1$ ms, flip angle = $12^{\circ}$, FOV = $24 \times 24$ cm, matrix = $256 \times 192$, slice thickness = $4$ mm, $73$ sec temporal resolution, $21$ dynamics). An intravenous bolus injection of $0.1$ mmol/kg of gadoterate meglumine (Gd-DOTA) was administered simultaneously. The total acquisition time for DCE-MRI was approximately $24$ minutes. Two pre-contrast acquisitions were carried out at flip angles of $2^{\circ}$ and $12^{\circ}$ to calculate pre-contrast longitudinal relaxation times.

\subsubsection{Preprocessing.}
Undersampling was retrospectively applied to the fully-sampled data in the $k_x$-$k_y$ plane using a randomized golden-angle sampling pattern \cite{Zhu2016} over time (see Fig.~\ref{fig:ResidualAndMask}-(b)) with a 10-fold undersampling factor. The pre-contrast first frame was fully sampled. Due to the low temporal resolution of our data, we estimated subject-specific vascular input functions (VIFs) extracted by averaging a few voxels located on the superior sagittal sinus where the inflow artefact was reduced compared to a feeding artery \cite{Heye2016}. Data augmentation was employed by applying rigid transformations on image slices. We generated random 2D+t undersampling masks to be applied on the images of different orientations. This allows the network to learn diverse patterns of aliasing artifacts. All the subject's data required for network training/testing were divided into non-overlapping 3D blocks of size $52 \times 52 \times 33$, resulting in 64 blocks per subject.

\subsubsection{Experimental setup.}
All experiments were performed in a leave-one-subject-out fashion. The networks were trained using the Adam optimizer with a learning rate of $10^{-3}$ (using a decay rate of $10^{-4}$) for 300 epochs and mini-batch size of 4.  
To demonstrate the advantage of the proposed method, we compare it with the state-of-the-art model-based iterative parameter reconstruction method using the MATLAB implementation provided by the authors \cite{Guo2017}. We use the concordance correlation coefficient (CCC) and structured similarity metric (SSIM) metrics to quantitatively assess the PK parameter reconstruction, and peak signal-to-noise ratio (PSNR) metric to assess the image reconstruction. 
Experiments were run on a NVIDIA GeForce Titan Xp GPU with 12 GB RAM.

\subsubsection{Results.}
\begin{figure}[t!]
 \centering
 \includegraphics[width=0.8\columnwidth]{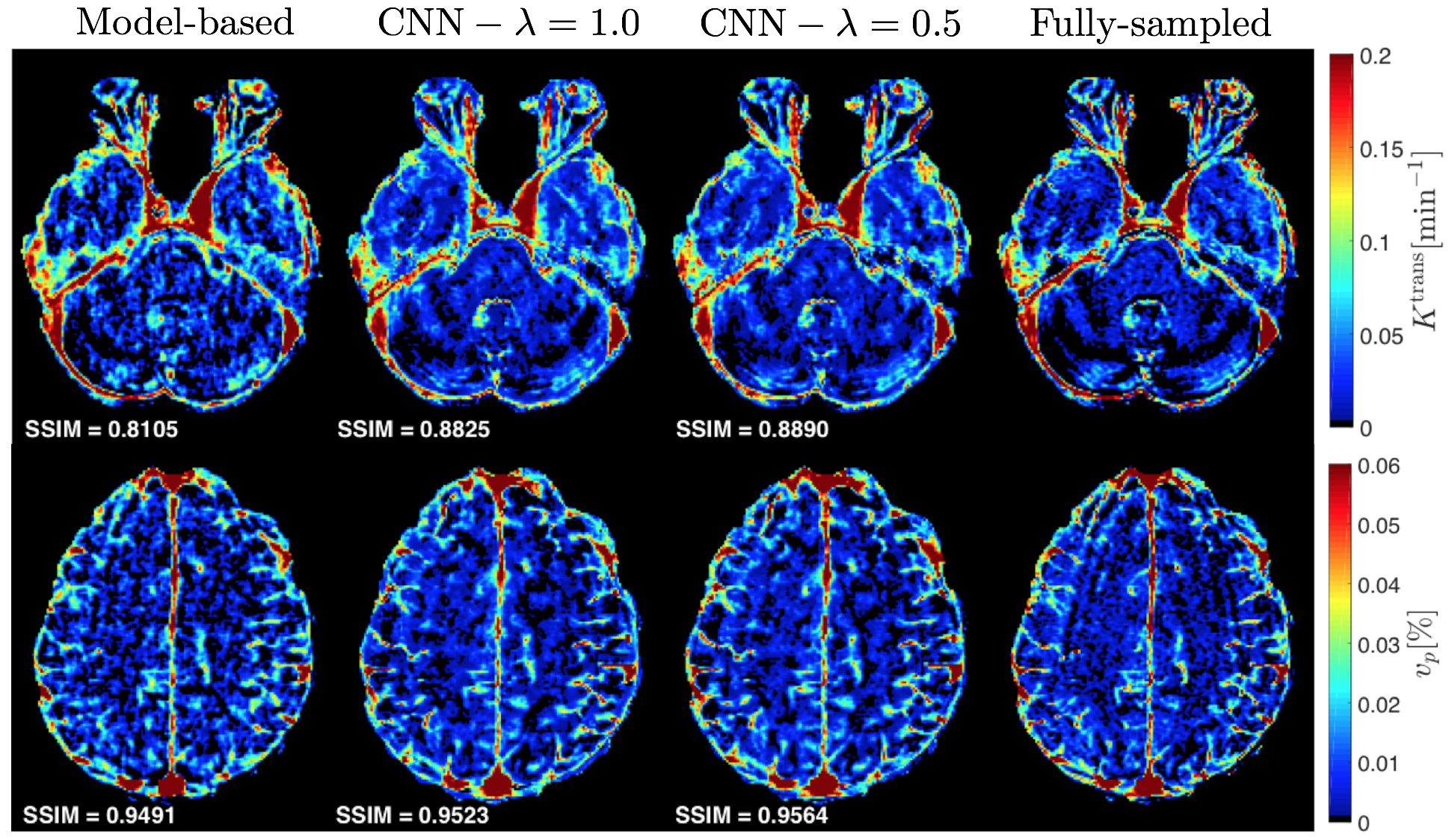}
 \caption{Reconstructed PK parameter maps of two exemplary slices of a test subject with a 10-fold undersampling. Brain masks are applied to estimated maps. Our CNN model incorporating both loss terms ($\lambda=0.5$) achieves the best paramater estimates. The resulting SSIM values are provided at the bottom-left corner of each map. \label{fig:ParamQualitative}}
\end{figure}

\begin{figure}[t!]
 \centering
 \includegraphics[width=0.85\columnwidth]{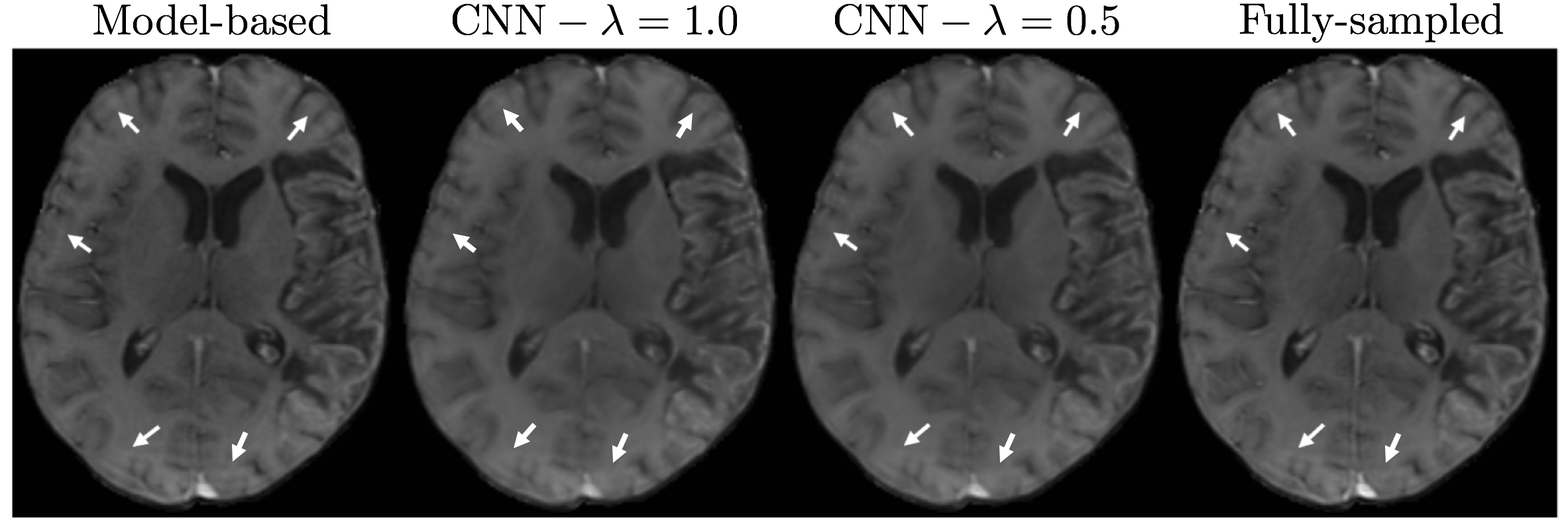}
 \caption{Visual comparison of the image reconstruction results of an examplary DCE slice. White arrows indicate a few regions where the Gibbs artifacts are observable. Our CNN model with both $\lambda=0.5$ and $1.0$ can significantly suppress the artifacts appearing in fully-sampled image and model-based reconstruction as well.  \label{fig:ImageQualitative}}
\end{figure}

\begin{figure}[t!]
\floatbox[{\capbeside\thisfloatsetup{capbesideposition={right,top},capbesidewidth=4.5cm}}]{figure}[\FBwidth]
{\caption{Parameter estimation (SSIM \& CCC) and image reconstruction (PSNR) performances calculated on all test slices for model-based (MB) reconstruction method and our proposed CNN model with different $\lambda$ settings. 
}\label{fig:QuantResults}}
{\includegraphics[width=7.5cm]{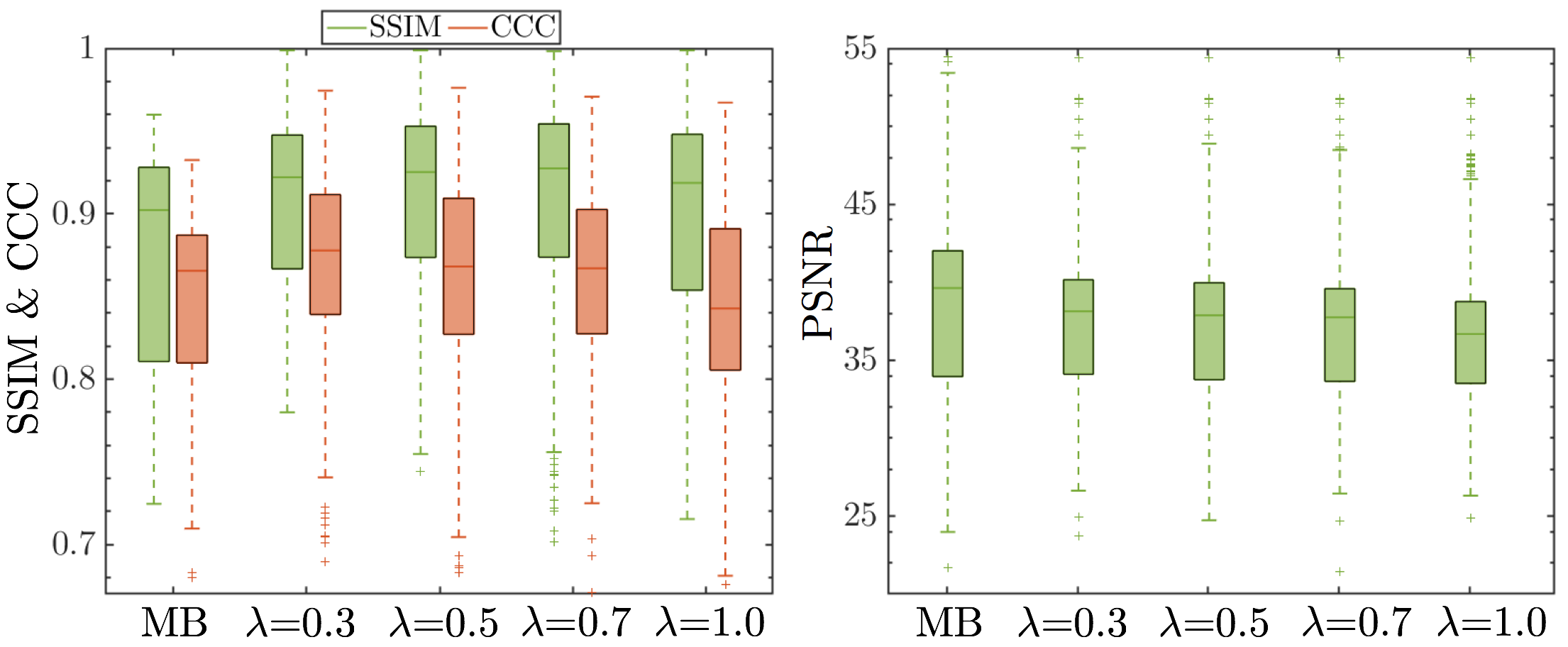}}
\end{figure}

Figure~\ref{fig:ParamQualitative} shows the qualitative PK parameter reconstructions obtained from different methods using 10-fold undersampling. The results indicate that CNN-$\lambda=0.5$ incorporating two loss terms simultaneously produces better maps and considerably higher SSIM score calculated with respect to fully-sampled PK maps. The model-based iterative reconstruction yields the PK maps where the artifacts caused by undersampling are still observable.
In Fig.~\ref{fig:ImageQualitative} we present the exemplary reconstructed images obtained by applying the operation $\boldsymbol{\tilde{f}_m}$ to the estimated PK maps. All the reconstruction approaches result in high quality images, however, the model-based reconstruction can better preserve the finer details. Unfortunately, our fully-sampled data suffer from Gibbs artifacts appearing as multiple parallel lines throughout the image. As marked by white arrows, our CNN method can significantly suppress these artifacts whereas they still appear in the image obtained by model-based iterative reconstruction. Finally, Fig.~\ref{fig:QuantResults} demonstrates the quantitative results of parameter estimation and image reconstruction. The highest CCC and SSIM values for parameter estimation are achieved by our CNN model when both loss terms are incorporated with $\lambda=0.3$ and $\lambda=0.5$, yielding an average score of 0.88 and 0.92, respectively. The difference is statistically significant for both CCC ($p=0.017$) and SSIM ($p=0.0086$) when compared against model-based reconstruction. The model-based reconstruction performs the highest PSNR for image reconstruction, where it is followed by the proposed CNN with $\lambda=0.3$. The difference between them is statistically significant with $p\ll0.05$. The PSNR also shows a decreasing trend with increasing $\lambda$ as expected. 


We emphasize that the parameter inference of our method on a 3D test volume takes around 1.5 seconds while the model-based method requires around 95 minutes to reconstruct the same volume, enabling $\approx 4\times 10^3$ faster computation.

\section{Conclusion}\label{sec:Conclusion}
\vspace{-2mm}
We present a novel deep learning based framework for direct estimation of PK parameter maps from undersampled DCE image-time series. Specifically, we design a \textit{forward physical model loss} function through which we exploit the physical model relating the contrast agent kinetics to the time-resolved DCE signals. Moreover,  we utilize the residual learning strategy in our problem formulation. The experiments demonstrate that our proposed method can outperform the state-of-the-art model-based reconstruction method, and allow almost instantaneous inference of the PK parameters in the clinical workflow of DCE-MRI. 


\subsubsection{Acknowledgements.}
The research leading to these results has received funding from the European Unions H2020 Framework Programme (H2020-MSCA-ITN- 2014) under grant agreement no 642685 MacSeNet. We also acknowledge Wellcome Trust (Grant 088134/Z/09/A) for recruitment and MRI scanning costs.

%
%
\bibliographystyle{splncs03}
\bibliography{references}

\end{document}